\documentclass[10pt,twocolumn,letterpaper]{article}

\usepackage[pagenumbers]{cvpr} 

\usepackage{multirow}
\usepackage{bbm}

\newcommand{\ourmethod}{GenMimic\xspace}
\newcommand{\ourbench}{GenMimicBench\xspace}

\definecolor{cvprblue}{rgb}{0.21,0.49,0.74}
\usepackage[pagebackref,breaklinks,colorlinks,allcolors=cvprblue]{hyperref}

\usepackage{xcolor}  

\definecolor{robotaction}{RGB}{255, 140, 0}  
\definecolor{robottype}{RGB}{178, 34, 34}
\definecolor{robottask}{RGB}{65, 105, 225}  
\definecolor{controltype}{RGB}{0. 205, 0}
\definecolor{predsteps}{RGB}{216, 191, 216}
\definecolor{lavendermist}{rgb}{0.9, 0.9, 0.98}
\definecolor{visualtrace}{RGB}{255, 216, 0} 
\usepackage{colortbl}
\usepackage{xcolor}
\definecolor{lightgray}{gray}{0.9}
\usepackage{booktabs}
\usepackage{color, soul}
\usepackage{graphicx}
\usepackage{amsmath}
\usepackage{empheq}
\usepackage{epigraph}
\usepackage{svg}
\definecolor{lightgray}{gray}{0.9}
\definecolor{lightblue}{rgb}{0.93,0.95,1.0}
\definecolor{darkgreen}{rgb}{0.0,0.6,0.0}
\definecolor{darkblue}{rgb}{0.0,0.0,0.5}
\definecolor{pinegreen}{rgb}{0.0, 0.47, 0.44}
\definecolor{deepmagenta}{rgb}{0.8, 0.0, 0.8}
\definecolor{amber}{rgb}{1.0, 0.49, 0.0}

\newcommand{\ignorebig}[1]{}

\def\Secref#1{Section~\ref{#1}}

\newcommand{\minisection}[1]{\noindent{\textbf{#1}.}}
\newcommand{\tabref}[1]{Table~\ref{#1}}

\newcommand{\figrref}[1]{Figure~\ref{#1}}

\newlength\savewidth

\definecolor{lightred}{RGB}{241,140,142}
\definecolor{amber(sae/ece)}{rgb}{1.0, 0.49, 0.0}
\definecolor{battleshipgrey}{rgb}{0.52, 0.52, 0.51}
\definecolor{cadmiumorange}{rgb}{0.93, 0.53, 0.18}
\definecolor{applegreen}{rgb}{0.55, 0.71, 0.0}
\definecolor{cadmiumgreen}{rgb}{0.0, 0.42, 0.24}
\definecolor{forestgreen}{rgb}{0.13, 0.55, 0.13}
\definecolor{red}{rgb}{0.89, 0.0, 0.13}

\definecolor{cb-0}{RGB}{216, 27, 96}
\definecolor{cb-1}{RGB}{30,136,229}
\definecolor{cb-2}{RGB}{255,193,7}
\definecolor{cb-3}{RGB}{0, 77, 64}
\definecolor{cb-4}{RGB}{150,220,174}

\title{From Generated Human Videos to Physically Plausible Robot Trajectories}

\author{James Ni* \textsuperscript{1}  
\quad Zekai Wang* \textsuperscript{1} 
\quad Wei Lin* \textsuperscript{3} 
\quad Amir Bar* \textsuperscript{1} 
\\
\quad Yann LeCun \textsuperscript{\textdagger} \textsuperscript{2} 
\quad Trevor Darrell \textsuperscript{\textdagger} \textsuperscript{1} 
\quad Jitendra Malik \textsuperscript{\textdagger} \textsuperscript{1} 
\quad Roei Herzig \textsuperscript{\textdagger} \textsuperscript{1} 
\\ \\
\textsuperscript{1} University of California, Berkeley \quad 
\textsuperscript{2} New York University \quad 
\textsuperscript{3} Johannes Kepler University
}

\newcommand\freefootnote[1]{%
  \let\svthefootnote\thefootnote
  \let\thefootnote\relax
  \footnotetext{#1}
  \let\thefootnote\svthefootnote
}

\begin{document}

\twocolumn[{%
\vspace{-1.5cm}
\renewcommand\twocolumn[1][]{#1}%
\maketitle
\begin{center}
    \centering
    \captionsetup{type=figure}
    \includegraphics[width=\linewidth]{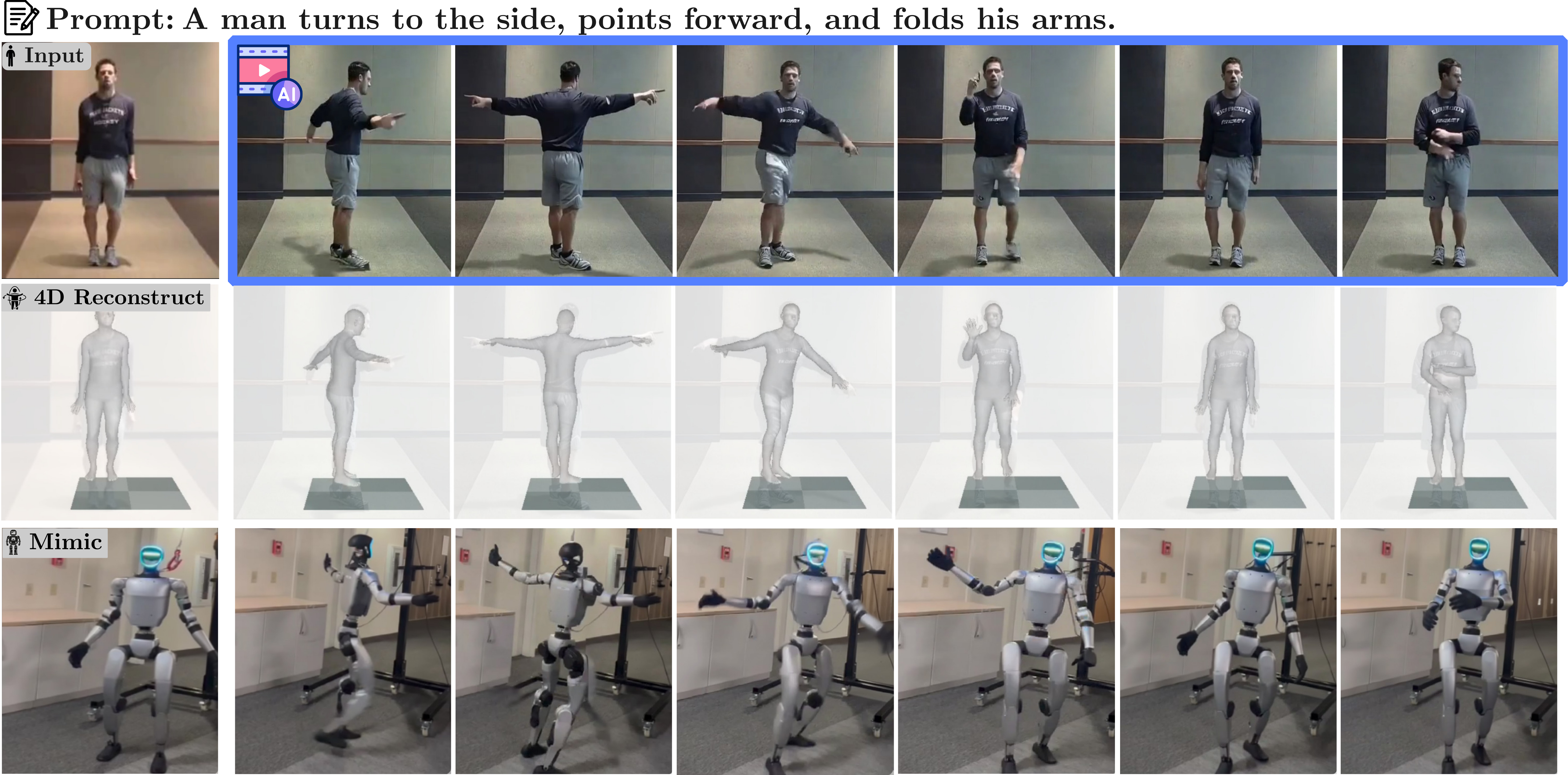}
    \captionof{figure}{\textbf{\ourmethod} enables zero-shot human control from generated videos. Given a text prompt (top) describing a human motion, a video generative model synthesizes a customized human-action video. The 4D human motion trajectory (mid) is reconstructed from the generated video, and mimicked by a Unitree G1 (bottom) using our robust policy, requiring no task-specific finetuning. The robot successfully reproduces the action sequence (\textit{turns, points, folds arms}) according to the synthetic video input in a zero-shot manner. Demonstration videos, code, checkpoints, and our dataset can be found on our website: \href{https://genmimic.github.io/}{https://genmimic.github.io/}.}
    \vspace{0.3cm}
\label{fig:teaser}
\end{center}
}]

\freefootnote{{*}Equal contribution.}
\freefootnote{{$\dagger$}Equal advising.}

\begin{abstract}
Video generation models are rapidly improving in their ability to synthesize human actions in novel contexts, holding the potential to serve as high-level planners for contextual robot control. To realize this potential, a key research question remains open: how can a humanoid execute the human actions from generated videos in a zero-shot manner?
This challenge arises because generated videos are often noisy and exhibit morphological distortions that make direct imitation difficult compared to real video.
To address this, we introduce a two-stage pipeline. First, we lift video pixels into a 4D human representation and then retarget to the humanoid morphology. Second, we propose \textbf{\ourmethod}—a physics-aware reinforcement learning policy conditioned on 3D keypoints, and trained with symmetry regularization and keypoint-weighted tracking rewards. As a result, \ourmethod can mimic human actions from noisy, generated videos.
We curate \ourbench, a synthetic human-motion dataset generated using two video generation models across a spectrum of actions and contexts, establishing a benchmark for assessing zero-shot generalization and policy robustness.
Extensive experiments demonstrate improvements over strong baselines in simulation and confirm coherent, physically stable motion tracking on a Unitree G1 humanoid robot without fine-tuning.
This work offers a promising path to realizing the potential of video generation models as high-level policies for robot control.
\end{abstract}
\section{Introduction}

Recent advances in humanoid control~\cite{videomimic, beyondmimic,h2o,omnih2o,gmt,humanplus,luo2023universal,perpetualcontrol} are driving progress toward general-purpose agents capable of performing human tasks. An unrealized requirement for such an agent is the ability to plan and adapt to unseen tasks and contexts.
At the same time, video generation models have emerged as powerful tools~\cite{hunyuanvideo, cosmospredict2, wan} for synthesizing behavior in novel contexts. As these models continue to improve in quality, they present a promising pathway for generative, vision-based planning and control. Realizing this potential raises a central research question: how can we enable a humanoid robot to faithfully reproduce the actions depicted in generated videos?

A key challenge with generated videos is that their inherent noise and morphological inaccuracies are too severe to use directly as training data.
Instead of training on generated videos, we aim to construct a robust policy capable of zero-shot mimicry.
To achieve this, our solution follows a two-stage pipeline. 
First, we employ existing 4D human reconstruction models, such as~\cite{tram,4dhumans}, to lift pixels to an intermediate human motion trajectory. To minimize the morphology gap from the first stage, we retarget the human motion trajectory to the robot morphology. Second, we train on AMASS~\cite{amass}, a physics-aware policy using reinforcement learning (RL) in IsaacGym~\cite{isaacgym}, conditioned on this retargeted representation, to predict the joint angles required to mimic the human action. See \figrref{fig:teaser} for an overview.

We introduce \ourmethod as the robust tracking policy required for this pipeline. The core insight is that certain keypoints are more important, and can be symmetrically correlated. Accordingly, \ourmethod employs two core features. First, we adapt the traditional tracking reward to use a weighted combination of 3D keypoints from the input motion, allowing the policy to selectively attend to the most important features. Second, we add an auxiliary symmetry loss to enforce learning the connections between symmetric keypoints. This loss introduces an inductive bias for robustness, providing an implicit mechanism for error correction: when one side of the motion is noisy or contains error, the policy can leverage information from its mirror reflection.

To systematically evaluate this capability, we construct a synthetic human-motion dataset \ourbench, generated using state-of-the-art video generation models: \textit{Wan2.1-VACE-14B}~\cite{wan} and \textit{Cosmos-Predict2-14B-Sample-GR00T-Dreams-GR1}~\cite{cosmospredict2}. 
\ourbench\ comprises {428} videos spanning controlled indoor and in-the-wild scenes. The Wan2.1 subset includes {217} clean, multi-view videos generated from NTU RGB+D~\cite{nturgbd} frames, covering five diverse subjects across structured action compositions and camera viewpoints. 
The Cosmos-Predict2 subset complements this with {211} videos generated from PennAction~\cite{pennaction} frames, featuring eight subjects performing simple and object-interaction motions in realistic environments.
Together, these partitions provide a diverse benchmark designed to probe humanoid policy robustness and zero-shot generalization across visual, morphological, and motion distribution shifts.

We summarize our main contributions as follows: (i) We present the the first generalist framework enabling humanoid robots to execute motions generated by video generation models. (ii) We introduce \ourmethod, a novel reinforcement learning policy trained with symmetry regularization and a selective weighted combination of 3D keypoint rewards, generalizing to noisy, synthetic videos despite being trained solely on existing motion capture data. (iii) We curate a synthetic human-motion dataset \ourbench using Wan2.1 and Cosmos-Predict2, establishing a scalable benchmark for assessing zero-shot generalization and policy robustness. (iv) We validate our approach extensively in simulation and real-world experiments. In simulation, we provide detailed ablations and demonstrate significant improvements over strong baselines. We further confirm our method's viability on a physical Unitree G1 robot, demonstrating coherent and physically stable motions.

\section{Related Work}

\minisection{Humanoid Policy Learning} Reward-based approaches for humanoid control have enabled humanoid robots to learn a wide range of locomotion skills \cite{nexttokenlocomotion, terrainlocomotion, denoisinglocomotion, lipschitzlocomotion, versatilebiped, rwlocomotion, unifiedcontroller}, contact-based skills \cite{wococo, gettingup, standingup}, and even parkour \cite{humanoidparkour}. However, these task-specific controllers rely on hand-crafted objectives, and do not easily generalize. 

In recent years, a complementary, data-driven learning paradigm using human motion capture data has emerged, driven by progress in physics-based simulation. Early works~\cite{deepmimic,ase} pioneered learning from motion for physically simulated characters. Strong progress in this domain has led to learning generalizable skills across a wide range of environments ~\cite{pdp,maskedmimic,freemotion,perpetualcontrol}. H2O~\cite{h2o}, OmniH2O~\cite{omnih2o}, and HumanPlus~\cite{humanplus} successfully extended motion tracking to robotics by learning whole-body controllers for humanoid teleoperation. 

Building off this strong progress, recent works including HOVER~\cite{hover}, GMT~\cite{gmt}, Any2Track~\cite{any2track}, BeyondMimic~\cite{beyondmimic}, and TWIST~\cite{twist} have demonstrated high-fidelity reproduction of diverse human motions. VideoMimic~\cite{videomimic} and ResMimic~\cite{resmimic} further extend these capabilities by introducing scene-aware and object-aware interaction and control, enabling robots to respond to environmental context. These approaches are driving progress toward general-purpose humanoid agents capable of performing diverse human tasks. Complementary to these works, we introduce a robust policy conditioned on 4D human reconstruction data that is capable of zero-shot mimicry and generalization to novel human motions generated by video generative models. This capability opens a pathway towards vision-based generative planning and control.


\minisection{Zero-shot Generalization for Robotics} Zero-shot generalization first emerged as a hallmark of large-scale language and vision-language models (VLMs).
CLIP~\cite{clip} demonstrated recognition and retrieval on unseen categories without fine-tuning, inspiring open-vocabulary detection and segmentation~\cite{vild,regionclip,ovseg,maskclip}.
Modern multimodal LLMs~\cite{flamingo,llavaov,gemini2.5,qwen2.5vl,internvl3,gemma3,gpt5,glm4.5v,ovis2.5,phi4} further extend this capability through in-context prompting, enabling broad zero-shot transfer across semantic and visual domains.
Building on this foundation, robotics research has extended zero-shot generalization from perception to action.
Vision-Language-Action (VLA) frameworks, such as RT-2~\cite{rt2}, SayCan~\cite{saycan}, CLIPort~\cite{cliport}, LLARVA~\cite{Niu2024LLARVAVI}, ARM4R~\cite{niu2025pretraining}, and OpenVLA~\cite{openvla}, ground robot control in multimodal representations to perform unseen tasks without retraining.
Concurrently, generalist humanoid policies~\cite{gmt,twist,beyondmimic} extend zero-shot behavior to full-body imitation and locomotion, but remain dependent on high-quality motion capture for input. 
In contrast, our work explores zero-shot imitation from motion reconstructed from noisy generated videos. \ourmethod\ enables humanoid policies to produce corresponding human actions directly from such video, without the need for fine-tuning.


\minisection{Video Generative Models for Human Motion Generation}
Recent progress in video generative models has enabled controllable human-motion synthesis with improved temporal coherence and semantic consistency.
Diffusion–transformer methods such as Wan~\cite{wan}, HunyuanVideo~\cite{hunyuanvideo}, and MovieGen~\cite{moviegen} capture structured, action-driven dynamics from text or visual prompts.
Identity-preserving approaches~\cite{phantom,idanimator,magref,hunyancustom,idfreq} ensure consistent subject appearance, while multimodal and multi-concept extensions~\cite{humo,interacthuman,conceptmaster,movieweaver} integrate text, image, and audio for fine-grained motion control.
Action-aware models like Cosmos-Predict2~\cite{cosmospredict2} and DreamGen~\cite{dreamgen} further link generative video synthesis with predictive modeling, producing physically plausible motion useful for embodied learning.
Building on this trend, \ourmethod\ investigates how a humanoid robot can execute human actions depicted in generated videos in a zero-shot manner.
We leverage Cosmos-Predict2 and Wan2.1 to produce diverse synthetic motions spanning varied subjects, viewpoints, and action compositions, reframing video generative models as potential action planners for robotic control.

\begin{figure*}[t]
    \centering
    \includegraphics[width=\linewidth]{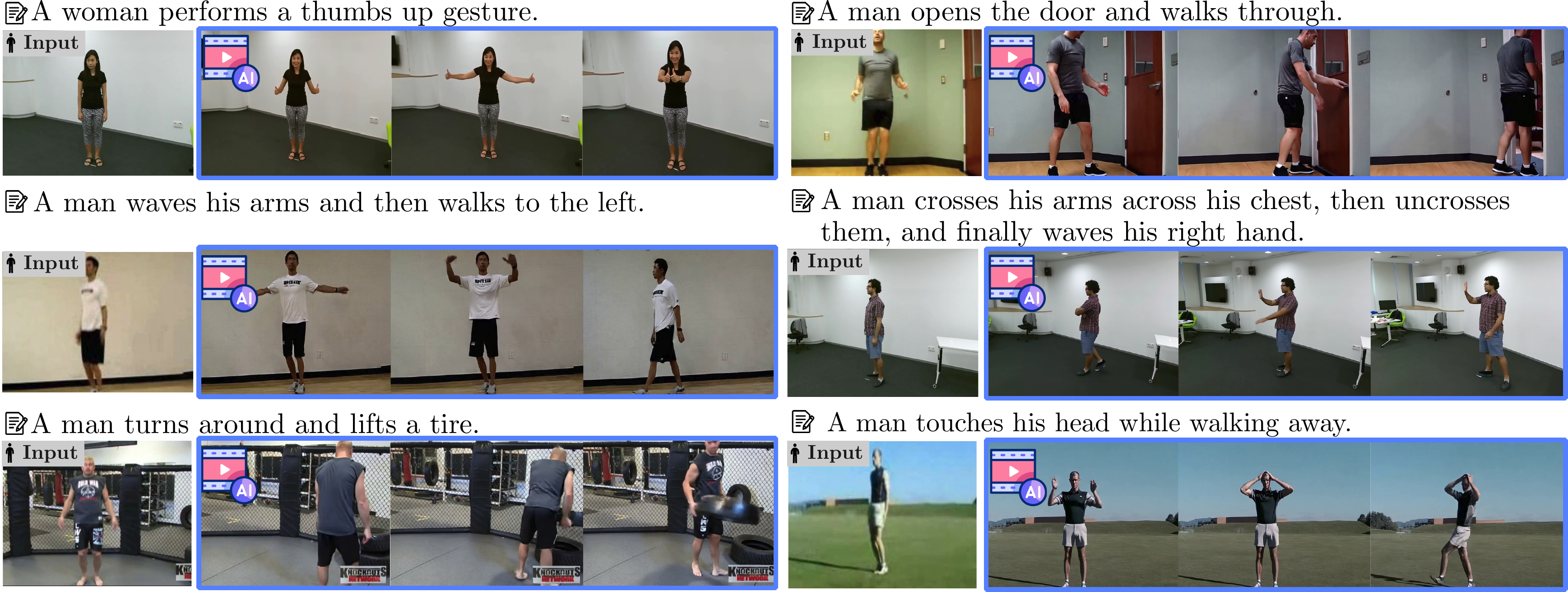}
    \caption{\textbf{Examples of \ourbench.} 
    Our synthetic human-motion dataset is generated using the Wan2.1 and Cosmos-Predict2 video generation models. These videos are produced conditioned on an initial frame and a text prompt describing the action. The dataset spans diverse subjects, environments, and action types including simple gestures and motion compositions.
    }
    \label{fig:dataset_figure}
\end{figure*}

\section{\ourbench}
To evaluate the zero-shot generalization of humanoid control policies under diverse visual and motion distributions, we introduce \ourbench, a synthetic human motion dataset comprising {428} generated videos. 
The dataset is created using two state-of-the-art video generation models, \textit{Wan2.1-VACE-14B}~\cite{wan} and \textit{Cosmos-Predict2-14B-Sample-GR00T-Dreams-GR1}~\cite{cosmospredict2}. 
As illustrated in~\figrref{fig:dataset_figure}, every sequence is generated from an initial frame and a text prompt specifying the intended action, enabling systematic variation in subject identity, viewpoint, and motion.
Overall, \ourbench spans a wide variety of subjects, environments, and action types, from simple gestures to multi-step compositions and object-interaction behaviors.

\minisection{Wan2.1 Videos: Controlled Indoor Scenes}
A large portion of \ourbench is generated from NTU RGB+D~\cite{nturgbd} frames using Wan2.1.
These clips provide clean, structured indoor environments with synchronized front, left, and right camera views.
We include five subjects with varied demographics, body proportions, and clothing styles, ensuring diversity in appearance while maintaining consistent scene geometry.
The motions span four structured categories:
(a) \textit{Simple Upper-Body Motions} (3 actions: touch head, thumbs up, wave arms); (b) \textit{Simple Upper-Body Motion + Walking} (4 actions: no upper-body motion, touch head, thumbs up, wave arms + walking); (c) \textit{Composite Upper-Body Motions} (4 sequences: touch head $\rightarrow$ thumbs up $\rightarrow$ wave arms, touch head $\rightarrow$ fold arms, raise right hand and point forward $\rightarrow$ fold arms, cross arms $\rightarrow$ uncross $\rightarrow$ wave right hand); and (d) \textit{Composite Upper-Body Motion + Walking} (4 sequences, combining the composite actions with walking). 
This yields \textit{217} multi-view indoor videos capturing fine-grained variations in morphology, viewpoint, and action composition.

\minisection{Cosmos-Predict2 Videos: Web-Style Scenes}
To complement these controlled scenes with greater diversity, 
we additionally generate videos conditioned on PennAction~\cite{pennaction} frames using Cosmos-Predict2.
These clips reflect in-the-wild YouTube video characteristics: cluttered scenes, varied camera motion, nonuniform lighting, and real-world object layouts.
The subset includes \textit{211} videos featuring eight distinct subjects performing both simple gestures (e.g., touch head, thumbs up) and a range of object-interaction behaviors, such as opening doors, lifting books or dumbbells, and manipulating everyday household items.
This partition exposes policies to realistic complexities absent in controlled datasets, providing a challenging test bed for evaluating robustness in natural environments.

In total, \ourbench provides a unified collection of \textit{428} high-variance synthetic motion sequences spanning structured indoor scenes and diverse real-world video contexts.
By integrating controlled actions with diverse in-the-wild human motions, \ourbench establishes a comprehensive benchmark for assessing zero-shot humanoid policy performance under visual, morphological, and motion distribution shifts.
The dataset is purposefully designed to stress-test robustness, making it well suited for evaluating policies that rely on noisy or imperfect motion reconstructions from generated videos.

\section{Generated Video to Humanoid Actions}

\begin{figure*}[t]
    \centering
    \includegraphics[width=\linewidth]{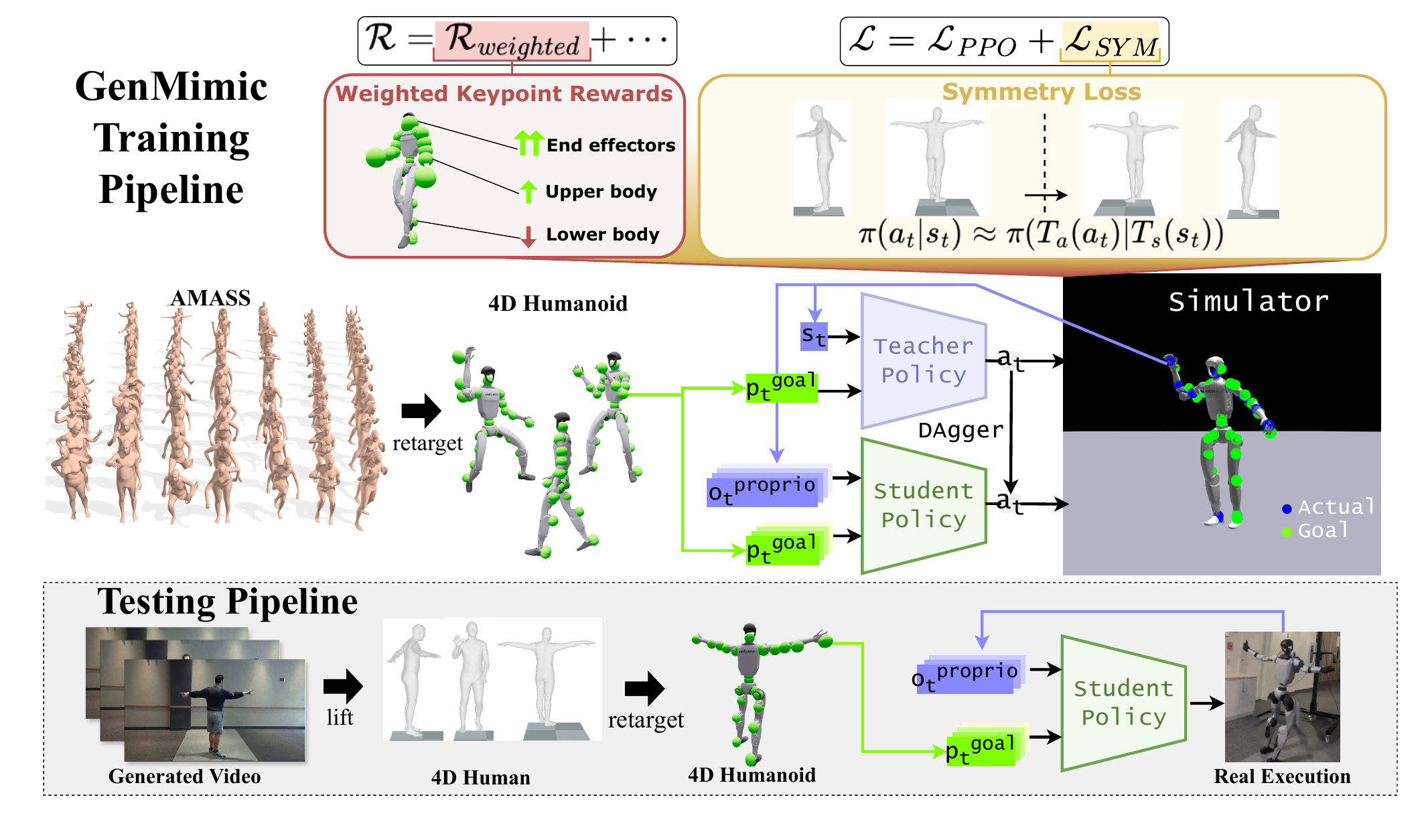}
    \caption{\textbf{\ourmethod Policy.} \textit{Training pipeline}: Our \ourmethod\ is trained in IsaacGym on retargeted AMASS trajectories via a student–teacher framework. Training incorporates two key features: (i) weighted keypoint rewards, which prioritize tracking of end effectors over the lower body; and (ii) a symmetry loss, which provides an inductive bias toward symmetric policies. \textit{Testing pipeline}: a generated video is lifted to a 4D human motion, and retargeted to the humanoid morphology and used as the goal input for \ourmethod. Finally, the policy executes the motion in the real world.}
    \label{fig:policy}
\end{figure*}

To address the challenge of executing humanoid actions from generated videos, we introduce a two-stage pipeline grounded in 4D reconstruction in Section \ref{sec:methods:pipeline}. We then describe in detail our novel, robust \ourmethod tracking policy in Section \ref{sec:methods:genmimic}. See \figrref{fig:policy} for an overview of our method.

\subsection{Two-Stage Pipeline}
\label{sec:methods:pipeline}

\minisection{Stage 1: From pixels to 4D humanoid reconstruction} Given a generated input RGB video, we use a state-of-the-art human reconstruction model to detect and extract the per-frame global pose ($\theta_t \in \mathbb{R}^{7}$) and SMPL \cite{smpl} parameters (shape $\beta_t \in \mathbb{R}^{16}$ and per-joint angle-axis $J_t \in \mathbb{R}^{J \times 3}$). Due to morphology mismatch, the resulting SMPL trajectory cannot be directly used for the humanoid. Hence, we retarget the SMPL trajectory to the robot's joint-space ($q_t^{\text{goal}}$), which combined with $\theta_t$ recovers the global 3D keypoints ($p_t^{\text{goal}}$) in robot space. 

\minisection{Stage 2: From 4D humanoid to actions} 
To properly generalize to unseen human actions, our policy must be robust to variations and noise in the input. To achieve this, we specifically choose 3D keypoints $p_t^{\text{goal}}$ over joint angles $q_t^{\text{goal}}$, as keypoints are more robust to variations and noise is more directly observable in this representation. 

Given these keypoints $p_t^{\text{goal}}$ and proprioceptive information ($o_t^{\text{proprio}}$), our tracking policy outputs physically-realizable desired joint angles ($q_t^{\text{des}}$). These desired joint angles are used by a proportional-derivative (PD) controller which outputs actionable torques to the robot.

Next, we describe in detail our robust tracking policy.

\subsection{The \ourmethod Policy}
\label{sec:methods:genmimic}

\subsubsection{Preliminaries}  
\label{sec:methods:prelim}

We formulate tracking humanoid motion as a decision problem modeled by a Partially Observable Markov Decision Process, defined by states $s_t \in \mathcal{S}$, observations $o_t \in \Omega$, actions $a_t \in \mathcal{A}$, and rewards $\mathcal{R}_t \in \mathbb{R}$. Since $s_t$ is difficult to estimate on the real robot, we train a privileged teacher policy ($\pi_s : \mathcal{S} \to \mathcal{A}$) in simulation using Proximal Policy Optimization (PPO) \cite{ppo}. We then distill the learned behaviors into a student policy ($\pi_o : \Omega \to \mathcal{A}$) using DAgger~\cite{dagger}.

At time $t$, we assume access to the robot's current proprioceptive state, which includes joint positions ($q_t$), joint velocities ($\dot{q}_t$), root angular velocity ($\omega_t^{\text{root}}$), projected gravity vector ($g_t$), and the immediate previous action ($a_{t-1}$). We additionally use the robot's 3D rigid body position in local space ($p_t^{\text{local}}$, with respect to the pelvis), calculated using forward kinematics. For the student policy, we concatenate the proprioceptive information ($o_t^{\text{proprio}} = [q_t, \dot{q}_t, \omega_t^{\text{root}}, p_t^{\text{local}}, g_t, a_{t-1}]$) of the past $\ell^{\text{proprio}}$ steps, along with the goal 3D keypoints of the $\ell^{\text{goal}}$ future steps.

The privileged teacher policy has access to the complete state information of the simulator. This information contains the robot's rigid body state information, including global 3D position ($p_t$), quaternion ($r_t$), linear velocity ($\dot{p}_t$), and angular velocity ($\omega_t$), as well as simulation parameters ($\theta^{\text{sim}}$). The teacher uses no history of proprioceptive information ($s_t = [o_t^{\text{proprio}}, p_t, r_t, \dot{p}_t, \omega_t, \theta^{\text{sim}}]$) or goal future, but also gets access to the difference between the goal and robot positions ($p_t^{\text{goal}} - p_t$).
For a complete table of the observation space, see Appendix \ref{supp:training:obstate}.

\begin{figure*}[th!]
    \centering
    \includegraphics[width=\linewidth]{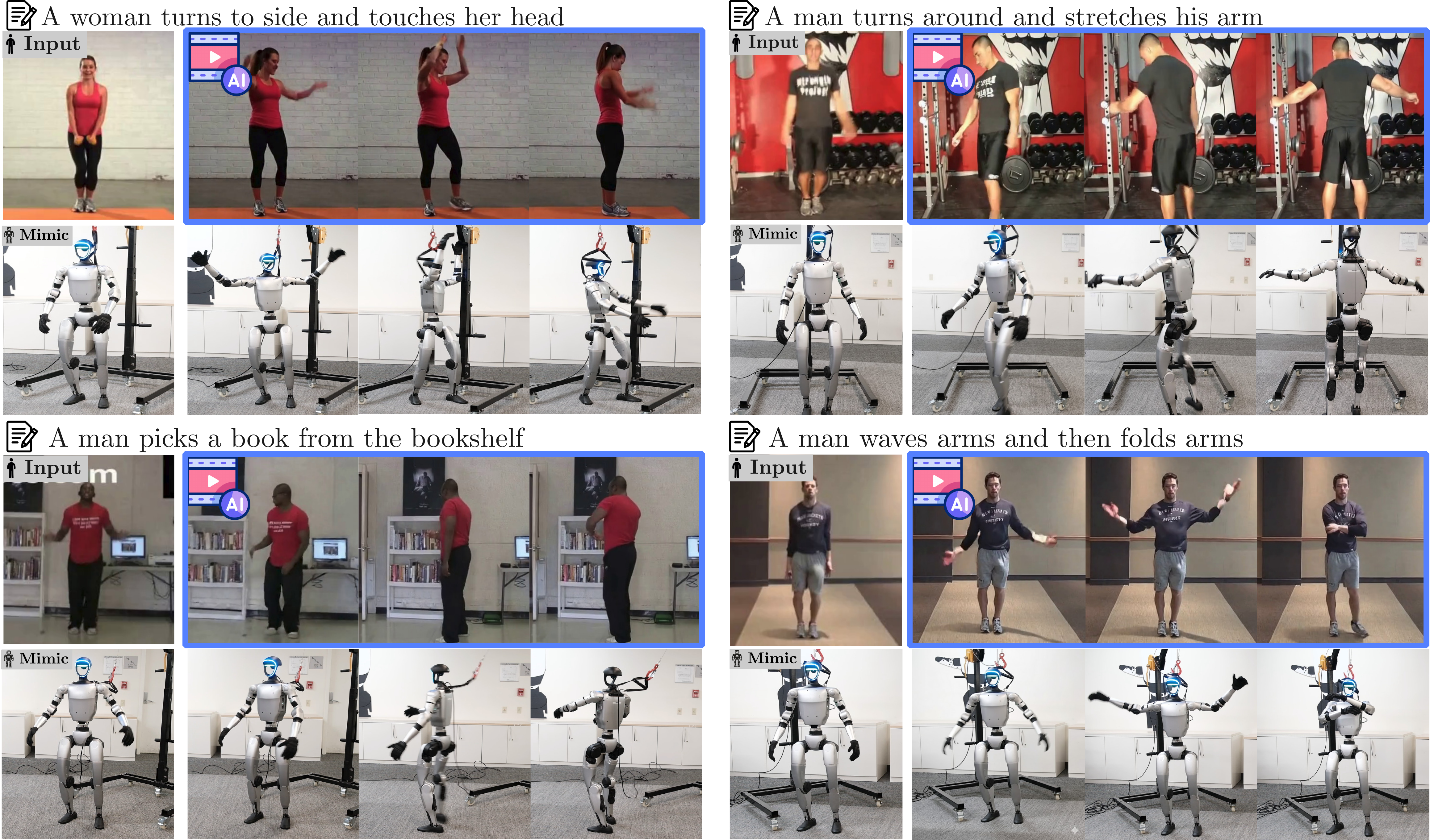}
    \caption{\textbf{Humanoid Mimic Examples.} Illustration examples ranging from simple gestures (e.g., \textit{touch head}) to composite and object-interaction motions (e.g., \textit{pick a book}). The visualization demonstrates that \ourmethod achieves coherent, physically plausible tracking across diverse actions and appearances. }
    \label{fig:mimic_result_demo}
\end{figure*}

\subsubsection{Robust Tracking Policy} 
\label{sec:methods:policy}

The tracking policy is illustrated in \figrref{fig:policy}. Generated human action from video contains noise and morphological mismatches that places them off-distribution from training data. We show that adding a weighted keypoint tracking reward and symmetry augmentation provides sufficient robustness to solve these challenges.

\minisection{Weighted Tracking} Certain keypoints, such as those corresponding to the end effectors, are inherently more critical for task execution and physical stability than torso or non-contact keypoints. Hence, we design a tracking reward to use a weighted combination of per-keypoint errors:
\begin{equation} 
\mathcal{R}_t^{\text{weighted}} = \exp\left(- \sum_{j=1}^{n} w_j \| p_{t,j} - p_{t,j}^{\text{goal}} \|_2^2 / \sigma^2 \right) 
\end{equation}
where each weight $w_j \geq 0$ and $\sum_{j=1}^n w_j = 1$. This formulation enables selective attention to the most reliable and task-relevant aspects of the goal. For generated video, a bias towards the end effectors and away from the inaccurate lower-body produces stable mimicry.

\minisection{Symmetry Loss} The human body exhibits inherent bilateral symmetry, where the left and right sides are approximate mirror images. We hypothesize that because this symmetry serves as a powerful physical inductive bias, a policy which explicitly learns and exploits these symmetric correlations between left and right keypoints can achieve greater robustness to per-keypoint noise in generated video. 

To accomplish this, we incorporate into the standard PPO training objective an auxiliary symmetry loss, $\mathcal{L}_{\text{SYM}}(\pi)$, controlled by a weighting coefficient $\lambda_{\text{SYM}}$:
\begin{equation}
\mathcal{L}_{\text{PPO-SYM}}(\pi) = \mathcal{L}_{\text{PPO}}(\pi) + \lambda _{\text{SYM}} \mathcal{L}_{\text{SYM}}(\pi)
\end{equation}

The symmetry loss ($\mathcal{L}_{\text{SYM}}$) is similar to the standard PPO loss, except that it uses a modified probability ratio. This modified ratio, $\beta_t^{\text{sym}}$, is calculated based on the symmetric state and action:
\begin{equation}
\beta_t^{\text{sym}} = \frac{\pi(T_a(a_t) | T_s(s_t))}{\pi_{\text{old}}(a_t | s_t)}
\end{equation}
where $T_s: S \to S$, $T_a: A \to A$ are the bilateral symmetry functions for states and actions. This loss effectively increases the likelihood of choosing the symmetric action $T_a(a_t)$ at the symmetric state $T_s(s_t)$ if the original action $a_t$ at state $s_t$ yields high advantage.

\subsubsection{Policy Learning}
\label{sec:methods:training}

In this section, we describe the policy learning details, including training data, rewards, and domain randomization.

\minisection{Data} We train \ourmethod on human motion trajectories from AMASS \cite{amass}, a large-scale motion capture dataset spanning 346 subjects and 11,451 motions across diverse activities like walking, running, dancing, and gesturing. We apply the retargeting process described in Stage 1 (\Secref{sec:methods:pipeline}) to retarget all human motion trajectories to the robot morphology. To ensure physical feasibility, we filter out trajectories involving object interaction or sufficiently high retargeting error, retaining 8,123 motions.

\minisection{Rewards} With our specialized tracking rewards and symmetry loss (\Secref{sec:methods:policy}), we find that a small number of reward function terms are sufficient for policy learning. We use four weighted tracking rewards based on the difference between the robot's joint angles, joint velocities, body positions, and body rotations, and the goal. We use three energy regularization terms based on acceleration, velocity, and action rate. We use two simulator-specific penalties to discourage undesired feet slippage and violation of joint limits. We use an orientation penalty to encourage stability. 

In addition to these core terms, we include the basic termination penalty and alive reward. We also include four feet penalties for safer stepping behavior in the real world. We note that all the non-weighted tracking terms are all standard reward terms for learning humanoid policies in RL.
Please see Appendix in \Secref{supp:training:rewards} for more details.

\minisection{Domain Randomization} To increase robustness, we apply input noise to the states and observations of our tracking policy. To ensure the policy is robust to external perturbations, environmental factors, and sim-to-real differences, we also randomize key simulation parameters. Specifically, we apply randomization to friction, mass, PD controller gains, motor strength, and control delay, and apply external impulses to the robot during simulation. For a detailed overview, please see Appendix in \Secref{supp:training:random}.

\begin{table*}[htpb]
\centering
\resizebox{2\columnwidth}{!}{%
\begin{tabular}{lccccc}
\toprule
& \multicolumn{5}{c}{\textbf{\ourbench}} \\
\cmidrule(lr){2-6}
\textbf{Method} \hspace{55px} & SR $\uparrow$ & MPKPE $\downarrow$ & LMPKPE $\downarrow$ & MPKPE-NT $\downarrow$ & LMPKPE-NT $\downarrow$ \\
\midrule
\multicolumn{6}{l}{\textit{Unprivileged Policies}} \\
\midrule
GMT & $4.29\%$ & $\mathbf{23.79 \pm 1.89}$ & $\mathbf{4.58 \pm 0.18}$ & $132.32 \pm 24.12$ & $\mathbf{6.78 \pm 0.79}$ \\
TWIST $(\pi_o)$ & $7.52\%$ & $24.16 \pm 2.33$ & $5.24 \pm 0.65$ & $98.08 \pm 57.09$ & $7.05 \pm 2.13$ \\
\ourmethod $(\pi_o)$ & $\mathbf{29.78\%}$ & $24.51 \pm 1.73$ & $5.98 \pm 0.42$ & $\mathbf{62.48 \pm 13.48}$ & $6.82 \pm 0.95$ \\
\midrule
\multicolumn{6}{l}{\textit{Privileged Policies}} \\
\midrule
TWIST $(\pi_s)$ & $2.69\%$ & $25.74 \pm 5.93$ & $\mathbf{6.61 \pm 1.88}$ & $120.96 \pm 59.46$ & $7.82 \pm 1.65$ \\ 
BeyondMimic & $23.81\%$ & $23.29 \pm 4.62$ & $10.19 \pm 6.47$ & $36.85 \pm 6.46$ & $\mathbf{5.58 \pm 0.38}$ \\
\ourmethod $(\pi_s)$ & $\mathbf{86.77}\%$ & $\mathbf{16.63 \pm 1.06}$ & $6.68 \pm 0.33$ & $\mathbf{20.46 \pm 5.73}$ & $6.75 \pm 0.47$ \\
\bottomrule
\end{tabular}
}
\caption{\textbf{Simulation Results}. Comparison of human motion tracking on \ourbench, separated by access to unprivileged and privileged information. $(\pi_o)$ denotes a student policy and $(\pi_s)$ denotes a teacher policy.}
\label{tab:experiments}
\end{table*}

\begin{table}[t!]
\centering
\resizebox{\columnwidth}{!}{%
\begin{tabular}{lcc|lcc}
\toprule
\textbf{Action Type} & \# & \textbf{VSR} $\uparrow$ & \textbf{Action Type} & \# & \textbf{VSR} $\uparrow$ \\
\midrule
\textbf{in-place action} & 12 & $1.0$ & \textbf{turn+action} & 12 & $0.41$ \\
\quad wave & 6 & $1.0$ & \quad $2 \times 180$ turn & 3 & $0.45$ \\
\quad reach & 2 & $1.0$ & \quad $2 \times 90$ turn & 3 & $0.44$ \\
\quad raise & 2 & $1.0$ & \quad $180$ turn & 3 & $0.33$ \\
\quad sequence & 2 & $1.0$ & \quad $90$ turn & 2 & $0.50$ \\
\textbf{step+action} & 15 & $0.40$ & \quad $<90$ turn & 1 & $1.0$ \\
\quad step forward & 7 & $0.43$ & \textbf{walk+action} & 4 & $0.60$ \\
\quad step lateral & 7 & $0.43$ & \quad walk & 2 & $1.0$ \\
\quad raise leg & 1 & $0.0$ & \quad walk+turn & 2 & $0.20$ \\
\bottomrule
\end{tabular}
}
\caption{\textbf{Real-world Results.} 
We deploy 43 total human actions from \ourbench on a Unitree G1 humanoid, separated into broad categories. For each category, we report the number of actions ($\#$) and the Visual Success Rate (VSR), measured across multiple (2-6) trials, averaged across all actions in that category.
}
\label{tab:realworld_deployment}
\end{table}

\section{Experiments}
We evaluate \ourmethod's performance in mimicking human actions in \Secref{sec:exp:results} using our policy trained on AMASS. This evaluation compares our method in simulation against strong baselines on our \ourbench dataset. We also evaluate our method in real setting with a 23-DoF Unitree G1 humanoid robot in \Secref{sec:exp:real}. Finally, we perform rigorous ablations showing our design choices in \Secref{sec:exp:ablations}. 

\subsection{Implementation Details}
\label{sec:exp:implementation}

The first stage of the pipeline utilizes TRAM \cite{tram} for 4D human reconstruction from video, followed by PHC \cite{perpetualcontrol} to retarget the 4D human motion to the robot embodiment. For the second stage, we implement \ourmethod in PyTorch as a Multi-layer Perceptron (MLP) with hidden layer dimensions of $[512, 256, 128]$. We implement symmetry loss similar to \cite{symmetry}. The tracking policy operates at $\mathbf{50\ \text{Hz}}$, while the PD controller runs at $\mathbf{200\ \text{Hz}}$ via synchronized $4\times$ sub-stepping. The student policy uses a proprioceptive history length of $\ell^{\text{proprio}} = 10$ and a goal future length of $\ell^{\text{goal}} = 10$. Training was conducted within IsaacGym~\cite{isaacgym} over $1.5B$ samples, utilizing four NVIDIA RTX 4090 GPUs. Deployment uses a single NVIDIA 4060 Mobile GPU. For further detials, see Appendix in \Secref{supp:training}.

\subsection{Baselines}
\label{sec:exp:baselines}

We compare \ourmethod to several existing state-of-the-art tracking policy baselines in simulation. GMT \cite{gmt} achieves general motion tracking using a mixture-of-experts teacher and adaptive sampling, and is conditioned on DoFs. TWIST \cite{twist} achieves high-quality performance by integrating real-world motion capture training data. BeyondMimic \cite{beyondmimic} proposes an alternative adaptive sampling strategy and a single set of generalizable hyperparameters, but relies on real-time state estimation. For privileged comparisons, we re-implement and train BeyondMimic on AMASS, and TWIST teacher on their publically released dataset. For unprivileged comparisons, we utilize GMT and TWIST student's publicly available pre-trained checkpoint. See Appendix in \Secref{supp:training:baselines} for more details.

\begin{table*}[ht!]
\centering
\resizebox{2\columnwidth}{!}{%
\begin{tabular}{lcccccccc}
\toprule
& \multicolumn{3}{c}{\textbf{AMASS}} & \multicolumn{5}{c}{\textbf{\ourbench}} \\
\cmidrule(lr){2-4} \cmidrule(lr){5-9}
\textbf{Method} \hspace{1.2in} & SR $\uparrow$ & MPKPE $\downarrow$ & LMPKPE $\downarrow$ & SR $\uparrow$ & MPKPE $\downarrow$ & LMPKPE $\downarrow$ & MPKPE-NT $\downarrow$ & LMPKPE-NT $\downarrow$ \\
\midrule
\multicolumn{6}{l}{\textit{Goal Observation Ablations}} \\
\midrule
DoFs & $45.8\%$ & $13.38 \pm 0.31$ & $6.65 \pm 0.09$ & $23.8\%$ & $25.64 \pm 1.56$ & $7.76 \pm 0.35$ & $53.84 \pm 6.67$ & $7.90 \pm 0.44$ \\
3D Points (3DP) & $50.0\%$ & $14.08 \pm 0.42$ & $7.19 \pm 0.15$ & $40.0\%$ & $23.23 \pm 1.69$ & $7.07\pm 0.55$ & $39.51 \pm 8.55$ & $7.22 \pm 0.70$ \\
\midrule
\multicolumn{6}{l}{\textit{Weight and Symmetry Ablations}} \\
\midrule
3DP+Weights & $97.7\%$ & $7.89 \pm 0.36$ & $6.09 \pm 0.14$ & $77.4\%$ & $\mathbf{16.59 \pm 1.39}$ & $7.03 \pm 0.50$ & $28.45 \pm 12.64$ & $7.77 \pm 1.30$ \\
3DP+Self-supervised & $67.6\%$ & $9.00 \pm 0.54$ & $6.10 \pm 0.17$ & $62.0\%$ & $18.63 \pm 1.32$ & $\mathbf{6.49 \pm 0.50}$ & $40.63 \pm 10.48$ & $7.75 \pm 0.99$ \\
\midrule
3DP+Weights+Symmetry & $\mathbf{99.3\%}$ & $\mathbf{7.49 \pm 0.36}$ & $\mathbf{5.62 \pm 0.09}$ & $\mathbf{86.8\%}$ & $16.63 \pm 1.06$ & $6.68 \pm 0.33$ & $\mathbf{20.46 \pm 5.73}$ & $\mathbf{6.75 \pm 0.47}$ \\
\bottomrule
\end{tabular}
}
\caption{\textbf{Ablations} on \ourmethod. We evaluate the impact of goal observation choice, weighted keypoint rewards, and symmetry loss on both AMASS (test) and \ourbench. We note that all policies are teachers, and (Success Rate) SR is the most reliable performance indicator, as the other metrics are inherently limited by the noise present in the generated videos used as ground truth.}
\label{tab:ablations}
\end{table*}

\subsection{Simulated Experiments}
\label{sec:exp:results}

We compared our method in simulation against strong baselines on \ourbench dataset. Results are in~\tabref{tab:experiments}.

To effectively compare policies, we report a range of metrics, averaged across 256 rollouts per motion from \ourbench. We define the Success Rate (SR) as the percentage of rollouts where the robot does not fall and its global position does not deviate by more than 0.5m from the goal. The policy's motion tracking fidelity is quantified by the mean per-keypoint positional error (cm) in both global (MPKPE) and local (LMPKPE) coordinates. Specifically, MPKPE measures overall tracking performance, while LMPKPE assesses the policy's ability to match the pose of the goal. 

However, standard MPKPE and LMPKPE metrics are biased towards successes, as they are calculated only over the rollout prior to termination. Consequently, comparing methods with varying success rates becomes misleading: a policy that fails early contributes errors only from its brief successful segments, whereas a robust policy accumulates errors over longer trajectories spanning broader, more challenging motions. To enable a unbiased comparison under the same input distribution, we additionally report the unconditional metrics MPKPE-NT and LMPKPE-NT (No Termination), which compute the metrics over an entire motion rollout without any termination condition. We note that all metrics provide useful signals, but SR remains the most critical, as the other metrics are more sensitive to noisy motion that serves as ground truth from generated videos.

As shown in \tabref{tab:experiments}, \ourmethod outperforms existing baselines in both privilieged and unprivilieged settings. The \ourmethod student achieves higher SR and MPKPE-NT than both GMT and TWIST, while the GenMimic teacher achieves higher SR, MPKPE, and MPKPE-NT than both BeyondMimic and TWIST. All unprivileged policies exhibit high global error, highlighting the challenge of zero-shot mimicking from generated video.

We attribute GMT's strong local pose tracking, but poor global tracking, to its reliance on DoF conditioning. Consequently, it fails to generalize off-distribution when faced with noisy motions. Both TWIST student and teacher exhibit high variance and poor performance across all metrics, presumably because they are designed to use high-quality motion capture as input. Finally, we note that BeyondMimic is most similar to the 3DP ablation in utilization of global positional information and observation space composition (but their rewards are quite different). While \ourmethod teacher also uses global information, we attribute its superior robustness on noisy motions to the inclusion of weighted keypoint rewards and the symmetry loss.

\subsection{Real-world Experiments}
\label{sec:exp:real}

We successfully deploy our policy on a 23-DoF G1 humanoid, demonstrating physical reproduction of human actions from generated video. We rollout 43 motions in total and report Visual Success Rate (VSR), in \tabref{tab:realworld_deployment}.
Unlike the quantitative simulation metrics that only measure deviation from ground truth, VSR evaluates whether the executed motion physically resembles the generated video. We consider any excessive stumbling or inability to visually follow a critical keypoint, such as a hand or foot, as a failure.

Our policy successfully reproduces a wide range of upper body motions, including waving, pointing, reaching, and sequences thereof. Composing these with lower body movements significantly increases difficulty. For stepping compositions, the policy reliably follows the upper body motion but fails to step or lift its leg consistently. For turning compositions, the policy reliably reaches the desired orientation but frequently stumbles. We hypothesize that these challenges stem from inaccurate or physically infeasible motion cues, a problem potentially solvable by introducing weighted noise to the 3D goal keypoints.

\subsection{Ablations}
\label{sec:exp:ablations}

We conduct ablations in simulation, trained on AMASS, to assess the importance of our three key design choices: the selection of goal observations, the use of weighted keypoint rewards, and the symmetry loss. Each ablation is trained on approximately 1.5B samples and evaluated on \ourbench and a 10\% test split of AMASS. We omit the NT metrics AMASS evaluation. The results are detailed in ~\tabref{tab:ablations}. More results are in Appendix in~\Secref{supp:add_ablations}.

\minisection{Goal Observation Ablations} We first ablate the goal observation space using a baseline policy that excludes the weighted keypoint reward and symmetry loss. We compare \textit{3D Points} (3DP) as detailed in \Secref{sec:methods:pipeline} and DoFs, which uses goal joint angles instead. Our experiments indicate that using 3D keypoints improves performance on noisy input.

\minisection{Weighted Keypoint Reward Ablations} Following the selection of \textit{3DP} as the optimal observation space, we next ablate the weighted keypoint reward. In \textit{3DP+Weights}, we defined a fixed weighting scheme to prioritize end effectors while de-prioritizing the lower body. Our experiments confirm the importance of these weights.

We also consider a data-driven, self-supervised approach to learning these weights. In the \textit{3DP+Self-supervised} configuration, the policy outputs an additional $n$-dimensional vector $\alpha_t$ at each time step $t$ with a regularization reward on changes to $\alpha_t$. Normalized weights are then computed via the Softmax function, $w_t = \text{Softmax}(\alpha_t)$. Our experiments show that learned weights achieve comparable performance to manually fine-tuned weights on noisy input, but worse performance on clean or difficult input.

\minisection{Symmetry Loss Ablation} We conclude our ablations by adding the symmetry loss to \textit{3DP+Weights} configuration. Note that the resulting \textit{3DP+Weights+Symmetry} is equivalent to \ourmethod teacher. Based on SR and the NT metrics, our experiments confirm that symmetry loss improves robustness to noise.

\section{Conclusion}
We present \ourmethod, a physics-aware humanoid control framework that enables robots to execute human motions depicted in generated videos in a zero-shot manner. We introduce a two-stage pipeline. First, we lift video pixels into a 4D human representation and then retarget to the humanoid morphology. Second, we propose \ourmethod, a physics-aware reinforcement learning policy conditioned on 3D keypoints, and trained with symmetry regularization and keypoint-weighted tracking rewards. We curate \ourbench, a scalable benchmark of synthetic human-motion videos generated using Wan2.1 and Cosmos-Predict2. We then evaluate our approach on this benchmark to assess its zero-shot generalization and policy robustness. Experiments in simulation and on a Unitree G1 humanoid demonstrate physically stable imitation and superior generalization compared to strong baselines. 

\section{Limitations and Future Work} 
\label{sec:limitations}

While our results demonstrate the feasibility of humanoids mimicking human motions to imitate generated videos, several limitations remain. First, the quality of motion trajectories is constrained by the quality of generated video and the downstream 4D reconstruction. Future work focusing on aligning the domain gap between generated and real video for reconstruction can enable a policy to be more robust and utilize richer scene information. Second, our policy is only trained on AMASS. We believe that the more diverse motion data a generalist policy can leverage, the better its performance will be on off-distribution motion. Third, our current evaluation primarily focuses on simple human actions rather than dynamic motions. Instead of conditioning directly on 3D keypoints, a promising direction is to learn on a latent motion representation that bridges simple, complex, real, and generated motions.
Ultimately, we believe general-purpose agents require the ability to plan and adapt to unseen tasks and contexts. This work serves as a first step toward that direction, paving the way for humanoids to engage in a vision-based generative planning and control.

\section*{Acknowledgments} We would like to thank Mahi Shafiullah, Justin Kerr, and Angjoo Kanazawa for helpful discussions and feedback. Authors, as part of their affiliation with UC Berkeley, were supported in part by the National Science Foundation, US Department of Defense, and/or the Berkeley Artificial Intelligence Research (BAIR) industrial alliance program, as well as the Humanoid Intelligence Center program. The views, opinions and/or findings expressed are those of the author and should not be interpreted as representing the official views or policies of any sponsor, the Department of Defense, or the U.S. Government.

{
    \small
    \bibliographystyle{ieeenat_fullname}
    \bibliography{main}
}


\newpage
\maketitlesupplementary
\newpage
\appendix

Here, we provide additional details on experiments and ablations (\Secref{supp:expr}), training procedures (\Secref{supp:training}), an analysis of \ourbench (\Secref{supp:our:datasets}), and the experimental hardware setup (\Secref{supp:g1}).

\section{Additional Experiment Results}
\label{supp:expr}

\subsection{Additional Ablations}
\label{supp:add_ablations}

We conduct additional ablations to asses the interplay of our design choices. Specifically, we investigate how the choice of goal observation can influence the effectiveness of weights and symmetry augmentation. Similar to \Secref{sec:exp:ablations}, we compare teacher policies trained on approximately 1.5B samples and evaluated on AMASS (test) and \ourbench. We report the same metrics and reemphasize SR as the most important metric. 

Results are in \tabref{tab:supp:ablations}. First, we find that conditioning on 3D keypoints improves performance on noisy inputs compared to conditioning on DoFs. Second, the use of weighted keypoint rewards consistently improves tracking fidelity regardless of the conditioning input. However, we note that improvement in robustness, especially to noisy inputs, is more pronounced when conditioning on 3D keypoints compared to DoFs. Finally, while the symmetry loss also improves robustness, this benefit is realized primarily when conditioning on 3D keypoints. This supports our hypothesis that the symmetry loss encourages the policy to learn the spatial relationship between the left and right sides, which is harder to learn in the DoF representation.

\begin{table*}[htbp]
\centering
\resizebox{2\columnwidth}{!}{%
\begin{tabular}{lcccccccc}
\toprule
& \multicolumn{3}{c}{\textbf{AMASS}} & \multicolumn{5}{c}{\textbf{\ourbench}} \\
\cmidrule(lr){2-4} \cmidrule(lr){5-9}
\textbf{Method} \hspace{1.1in} & SR $\uparrow$ & MPKPE $\downarrow$ & LMPKPE $\downarrow$ & SR $\uparrow$ & MPKPE $\downarrow$ & LMPKPE $\downarrow$ & MPKPE-NT $\downarrow$ & LMPKPE-NT $\downarrow$ \\
\midrule
DoFs & $45.8\%$ & $13.38 \pm 0.31$ & $6.65 \pm 0.09$ & $23.8\%$ & $25.64 \pm 1.56$ & $7.76 \pm 0.35$ & $53.84 \pm 6.67$ & $7.90 \pm 0.44$ \\
3D Points (3DP) & $50.0\%$ & $14.08 \pm 0.42$ & $7.19 \pm 0.15$ & $40.0\%$ & $23.23 \pm 1.69$ & $7.07\pm 0.55$ & $39.51 \pm 8.55$ & $7.22 \pm 0.70$ \\
DOFs+Weights & $66.9\%$ & $10.68 \pm 0.85$ & $6.68 \pm 0.23$ & $40.4\%$ & $17.61 \pm 1.71$ & $7.37 \pm 0.47$ & $58.20 \pm 18.09$ & $11.05 \pm 2.71$ \\
3DP+Weights & $97.7\%$ & $7.89 \pm 0.36$ & $6.09 \pm 0.14$ & $77.4\%$ & $\mathbf{16.59 \pm 1.39}$ & $7.03 \pm 0.50$ & $28.45 \pm 12.64$ & $7.77 \pm 1.30$ \\
DOFs+Weights+Symmetry & $84.0\%$ & $9.28 \pm 0.62$ & $\mathbf{5.53 \pm 0.16}$ & $40.0\%$ & $18.26 \pm 1.91$ & $6.73 \pm 0.44$ & $52.13 \pm 11.00$ & $7.95 \pm 0.95$ \\
3DP+Weights+Symmetry & $\mathbf{99.3\%}$ & $\mathbf{7.49 \pm 0.36}$ & $5.62 \pm 0.09$ & $\mathbf{86.8\%}$ & $16.63 \pm 1.06$ & $\mathbf{6.68 \pm 0.33}$ & $\mathbf{20.46 \pm 5.73}$ & $\mathbf{6.75 \pm 0.47}$ \\
\bottomrule
\end{tabular}
}
\caption{\textbf{Additional Ablations} on \ourmethod. We note that all polices are teachers. SR is the percentage of rollouts where the robot does not fall and stays within 0.5m of the goal. MPKPE and LMPKPE represent the mean per-keypoint positional errors in global and local coordinates prior to termination, respectively. We also report the unbiased metrics MPKPE-NT and LMPKPE-NT, which are computed over the full motion duration without termination.}
\label{tab:supp:ablations}
\end{table*}

\section{Additional Training Details}
\label{supp:training}

We next provide more details on Implementation Details, Observation and State Space, Rewards, Domain Randomization, and Baselines.

\subsection{Implementation Details}
\label{supp:impl}

We train the teacher policy using the PPO algorithm as implemented by RSL-RL~\cite{rslrl}. The policy operates at $50$ Hz, and we simulate physics 4 times per step, using a PD controller and IsaacGym at $200$ Hz. PPO hyperparameters are detailed in \tabref{tab:supp:ppo}. 

For the student policy, instead of employing a modified loss function, we apply symmetry augmentation to the training batch and directly minimize the $\ell_2$ loss between the student and teacher actions. 

To ensure a desirable distribution of actions, we clip the actions $a_t$ output by the actor policy and map them to the desired joint angles $q_t^{\text{des}}$ according to the following equation:
\begin{equation}
q_t^{\text{des}} = q^{\text{default}} + c \cdot \text{clip}(a_t, -a^{\text{clip}}, a^{\text{clip}})
\end{equation}
where $q^{\text{default}}$ is the default joint position, $c = 0.25$ is an action scaling parameter, and $a^{\text{clip}} = 10 \text{ rad}$ defines the clipping magnitude. These desired joint angles are then passed to a PD controller, which outputs the torques
\begin{equation}
T_t = K_p (q_t^{\text{des}} - q_t) - K_d \dot{q}_t
\end{equation}

During training, we terminate the episode if the robot's average keypoint position deviates more than $0.5$m from the goal keypoints, or if the projected gravity vector's $x$ or $y$ component exceeds 0.7 $\text{m}/\text{s}^2$. If the policy reaches the end of a motion, we resample the motion without resetting the environment, so that the policy can learn transition behavior between different motions.

\begin{table}
\centering
\begin{tabular}{l r}
\toprule
\textbf{Parameter} & \textbf{Value} \\
\midrule
\midrule
\multicolumn{2}{l}{\textit{PPO}} \\
\midrule
Number of GPUs & 4 RTX 4090's \\
Number of Environments & 4096 \\
Learning Epochs & 5 \\
Steps per Environment & 24 \\
Minibatch Size & 24576 \\
Discount ($\gamma$) & 0.99 \\
GAE ($\lambda$) & 0.95 \\
PPO Clipping Parameter & 0.2 \\
Entropy Loss Coefficient & 0.005 \\
Optimizer & Adam \\
Learning Rate & 1e-3 \\
Learning Rate Schedule & adaptive \\
Desired KL & 0.01 \\
Normalize Input & True \\ 
Normalize Value & False \\
\midrule
\multicolumn{2}{l}{\textit{DAgger}} \\
\midrule
Number of GPUs & 1 RTX 4090 \\
Number of Environments & 2048 \\
Learning Epochs & 5 \\
Optimizer & Adam \\
Learning Rate & 1e-3 \\
\bottomrule
\end{tabular}
\caption{\textbf{Hyperparameters of PPO and DAgger}.}
\label{tab:supp:ppo}
\end{table}

\subsection{Observation and State Space}
\label{supp:training:obstate}

\tabref{tab:supp:observations} details the policy's proprioceptive history and motion future, as well as the state space used by the privileged teacher policy and critics. In order to improve tracking quality, we extend the robot kinematic structure to include rigid body information for the head and hands. These are present and are the most relevant keypoints on the physical robot, but are not included in the standard kinematic chain. 

\begin{table}[t]
\centering
\resizebox{\columnwidth}{!}{%
\begin{tabular}{l|cccc}
\toprule
\textbf{Input} & \textbf{Dim.} & \textbf{Actor ($\pi_s$)} & \textbf{Actor ($\pi_o$)} & \textbf{Critic} \\
\midrule
\multicolumn{5}{l}{\textit{Proprioceptive History}} \\
\midrule
Joint positions & 23 & \checkmark & \checkmark & \checkmark \\
Joint velocities & 23 & \checkmark & \checkmark & \checkmark \\
Root angular velocity & 3 & \checkmark & \checkmark & \checkmark \\
Projected gravity & 3 & \checkmark & \checkmark & \checkmark \\
Previous action & 23 & \checkmark & \checkmark & \checkmark \\
Local rigid body pos. & $3 \times 27$ & \checkmark & \checkmark & \checkmark \\
Global rigid body pos. & $3 \times 27$ & \checkmark & $\times$ & \checkmark \\
Global rigid body quat & $4 \times 27$ & \checkmark & $\times$ & \checkmark \\
Global rigid body lin. vel. & $3 \times 27$ & \checkmark & $\times$ & \checkmark \\
Global rigid body ang. vel. & $3 \times 27$ & \checkmark & $\times$ & \checkmark \\
\midrule
History length & & $1$ & $10$ & $1$ \\
\textbf{Proprioceptive total} & & \textbf{507} & \textbf{1560} & \textbf{507} \\
\midrule
\multicolumn{5}{l}{\textit{Goal Futures}} \\
\midrule
Goal 3D keypoint pos. & $3 \times 27$ & \checkmark & \checkmark & \checkmark \\
Global pos. diff. & $3 \times 27$ & \checkmark & $\times$ & \checkmark \\
\midrule
Future length & & $1$ & $10$ & $1$ \\
\textbf{Goal total} & & \textbf{162} & \textbf{810} & \textbf{162} \\
\midrule
\multicolumn{5}{l}{\textit{Privilieged Information}} \\
\midrule
Base center of mass bias & 3 & \checkmark & $\times$ & \checkmark \\
Feet friction & 2 & \checkmark & $\times$ & \checkmark \\
Randomized mass & 8 & \checkmark & $\times$ & \checkmark \\
KD scale & 23 & \checkmark & $\times$ & \checkmark \\
KP scale & 23 & \checkmark & $\times$ & \checkmark \\
Torque scale & 23 & \checkmark & $\times$ & \checkmark \\
Feet contact forces & $2 \times 3$ & \checkmark & $\times$ & \checkmark \\
\textbf{Privileged total} & & \textbf{88} & \textbf{0} & \textbf{88} \\
\midrule
\midrule
\textbf{Total observation space} & & \textbf{676} & \textbf{2370} & \textbf{676} \\
\bottomrule
\end{tabular}
}
\caption{\textbf{Observations.}}
\label{tab:supp:observations}
\end{table}

\subsection{Rewards} 
\label{supp:training:rewards}

A full breakdown of the reward terms and their weights is provided in \tabref{tab:supp:rewards}. We highlight ten reward terms critical to enable the policy to mimic the input motion.  based on the weighted reward formulation detailed in \Secref{sec:methods:policy}.

We use four weighted rewards based on the formulation detailed in \Secref{sec:methods:policy}. In addition to keypoint position, we generalize our formulation to track joint position, joint velocity, and keypoint orientation. For weighted joint rewards, we use weights $w^{\text{upper}} = 2$ and $w^{\text{lower}} = 1$. For weighted keypoint rewards, we treat the hand and head keypoints as end effectors, with weights $w^{\text{end-eff}} = 4$, $w^{\text{upper}} = 2$, and $w^{\text{lower}} = 1$. 

We also utilize three regularizers based on physically and biologically grounded principles (such as energy minimization) to encourage desirable policy behavior. Due to simulator inaccuracies, we include penalties for exceeding joint limits and feet slipping behavior. To further enforce stability, we include an orientation reward.

In addition to these core terms, we also include a termination penalty and an alive reward. Finally, we also include four additional feet penalties to create safer stepping behavior in real and prevent hardware damage. We note that all the non-weighted tracking terms are all standard reward terms for learning humanoid policies in RL.

\begin{table*}[t]
\centering
\resizebox{2\columnwidth}{!}{%
\begin{tabular}{l|l|c|l}
\toprule
\textbf{Reward} & \textbf{Details} & \textbf{Weight} & \textbf{Rationale} \\
\midrule
\midrule
\multicolumn{4}{l}{\textit{Tracking Rewards}} \\
\midrule
Tracking Joint Pos & $\exp \left( \sum_j w_j (q_{t,j} - q_{t,j}^{\text{goal}})^2 \right) / \sigma_{\text{jp}}^2$ & 32 & \multirow{2}{*}{\shortstack[l]{while not physically feasible, the retargeted goal \\ joint angles are a good signal for correct behavior}} \\
Tracking Joint Vel & $\exp \left( \sum_j w_j (\dot{q}_{t,j} - \dot{q}_{t,j}^{\text{goal}})^2 \right) / \sigma_{\text{jv}}^2$ & 16 \\
Tracking Body Pos & $\exp \left( \sum_j w_j ||p_{t,j} - p_{t,j}^{\text{goal}}||^2 \right) / \sigma_{\text{bp}}^2$ & 50 & \multirow{2}{*}{\shortstack[l]{encourage proper mimicry of the goal motion by \\ prioritizing the tracking of 3D keypoints}} \\
Tracking Body Rot & $\exp \left( \sum_j w_j d_\text{quat}(r_{t,j}, r_{t,j}^{\text{goal}})^2 \right) / \sigma_{\text{br}}^2$ & 20 \\
\midrule
\multicolumn{4}{l}{\textit{Penalties / Regularization}} \\
\midrule
Action Rate & $\|a_{t-1} - a_t\|^2$ & $-1$ & slower behavior reduces sim-to-real gap \\
Energy & $\|T_t \odot q_t\|^2$ & $-1e-6$ & minimize effort applied \\
DoF Acceleration & $\|\ddot{q}_t\|^2$ & $-3e-6$ & discourage jittery movements \\
DoF Limits & $\sum \mathbbm{1}[(q_t > q^{ul}) \lor (q_t < q^{ll})]$ & $-100$ & penalize joints past the hard limits \\
Feet Slip & $\dot{p}_t^{\text{foot}} \cdot \mathbbm{1}[||F_t^{\text{foot}}|| > 1]$ & $-5$ & prevent feet slippage (simulator behavior) \\
Orientation & $||g_t^{xy}||^2$ & $-50$ & encourage stable orientation \\
\midrule
\midrule
\multicolumn{4}{l}{\textit{Feet Penalties}} \\
\midrule
Feet Contact Rewards & $\sum ||F_t^{\text{foot}}||$ & $-0.03$ & encourages less heavy steps \\
Feet Orientation & $\sum ||g_t^{\text{foot}, xy}||^2$ & $-62.5$ & encourages straight feet \\
Feet Max Height & $\sum \min (h_{\text{air}} - h_{\text{air}}^{\text{des}}, 0) \cdot \mathbbm{1}[\text{feet in air}]$ & $-2500$ & discourages steps too high \\
Feet Air Time & $\sum (t_{\text{air}} - t_{\text{air}}^{\text{des}}) \cdot \mathbbm{1}[\text{first step}] $ & $1000$ & encourages longer steps \\
\midrule
\multicolumn{4}{l}{\textit{Episodic Rewards}} \\
\midrule
Termination & on early termination & $-200$ & \\
Alive & on environment step & $20$ & \\
\bottomrule
\end{tabular}
}
\caption{\textbf{Rewards.} The upper half denotes the reward terms highlighted and crucial to \ourmethod. The lower half denotes the episodic rewards and feet penalties for safe deployment in real. $d_\text{quat}(r_1, r_2)$ is a distance function between quaternions. $q^{ul}$ and $q^{ll}$ denote the lower and upper limits of the robot's joints. $F_t^{\text{foot}}$ denotes the 3D contact forces at the feet rigid bodies. $g_t^{\text{foot}}$ denotes the 3D projected gravity vector based on the feet rigid body quaternion. $h_{\text{air}}$ and $h_{\text{air}}^{\text{des}} = 0.1$m denote the current maximum and desired maximum heights for the feet for a single step. $t_{\text{air}}$ and $t_{\text{air}}^{\text{des}} = 0.25$s denote the current and desired time in the air for the feet for a single step.}
\label{tab:supp:rewards}
\end{table*}

\subsection{Domain Randomization} 
\label{supp:training:random}

We perform extensive domain randomization in simulation to improve policy robustness and facilitate sim-to-real transfer. We apply random impulses to the robot every 5s to improve robustness to external disturbances. Upon environment reset, we randomize fundamental properties about the environment, robot, joints, and PD controller, ensuring the policy can mimic across diverse surfaces and physical conditions. Finally, we inject observation noise for the actors only, to ensure robustness to both real-world sensor noise, and noise from generated video. For a detailed breakdown, see \tabref{tab:supp:domain_rand}.

\begin{table*}[t]
\centering
\resizebox{2\columnwidth}{!}{%
\begin{tabular}{l|l|l|l}
\toprule
\textbf{Parameter} & \textbf{Type} & \textbf{Range} & \textbf{Rationale} \\
\midrule
\multicolumn{4}{l}{\textit{Perturbations}} \\
\midrule
Random push strength ($xy$) & Set Value & $\mathcal{U}[-1, 1]$ m/s & Robustness to impulses \\
\midrule
\multicolumn{4}{l}{\textit{Reset}} \\
\midrule
Ground friction & Set Value & $\mathcal{U}[0.4, 1.25]$ & Robustness to surface type \\
Base CoM & Additive & $\mathcal{U}[-100, 100]$ g & Robustness to weight imbalance \\
Link mass & Multiplicative & $\mathcal{U}[0.7, 1.3]$ & Robustness to weight imbalance \\
P gains ($K_p$) & Multiplicative & $\mathcal{U}[0.75, 1.25]$ & Robustness to controller error \\
D gains ($K_d$) & Multiplicative & $\mathcal{U}[0.75, 1.25]$ & Robustness to controller error \\
Motor strength & Multiplicative & $\mathcal{U}[0.5, 1.5]$ & Robustness to battery power, motor wear \\
Control delay & Set Value & $\{0, 1, 2, 3\}$ steps & Robustness to real-world input delay \\
Goal 3D keypoint offset & Additive & $\mathcal{U}[-0.02, 0.02]$ m & Robustness to goal input drift \\
\midrule
\multicolumn{4}{l}{\textit{Noise}} \\
\midrule
Joint position & \multirow{10}{*}{Additive} & $\mathcal{N}(0, 0.01)$ rad & \multirow{9}{*}{Robustness to real-world sensor noise}\\
Joint velocity & & $\mathcal{N}(0, 0.1)$ rad/s & \\
Root angular velocity & & $\mathcal{N}(0, 0.5)$ rad/s& \\
Projected gravity & & $\mathcal{N}(0, 0.1)$ m/s$^2$ & \\
Local rigid body position & & $\mathcal{N}(0, 0.01)$ m & \\
Global rigid body position & & $\mathcal{N}(0, 0.01)$ m & \\
Global rigid body quaternion & & $\mathcal{N}(0, 0.01)$ & \\
Global rigid body linear velocity & & $\mathcal{N}(0, 0.2)$ m/s & \\
Global rigid body angular velocity & & $\mathcal{N}(0, 0.5)$ rad/s & \\
Goal 3D keypoint positions & & $\mathcal{N}(0, 0.05)$ m & Robustness to goal input noise \\
\bottomrule
\end{tabular}
}
\caption{\textbf{Domain Randomization.} $\mathcal{U}$ denotes a uniform distribution and $\mathcal{N}$ denotes a normal distribution.}
\label{tab:supp:domain_rand}
\end{table*}

\subsection{Baselines} 
\label{supp:training:baselines}

\minisection{GMT} GMT achieves general motion tracking using the standard student-teacher framework, but with a mixture-of-experts teacher and adaptive sampling. Unlike our method, GMT conditions on DoFs and goal linear velocity, and does not use any positional information. We use GMT's pretrained checkpoint, implemented in a parallelized IsaacGym environment for fast evaluation.

\minisection{TWIST} TWIST similarly adopts the student-teacher framework, but instead of using DAgger, trains the student policy using RL, augmented with a KL-divergence loss between the student and teacher. This distillation procedure produces a student which performs better than the teacher. For the unprivileged student, we use TWIST's pretrained checkpoint, and similarly implement a parallelized IsaacGym environment for fast evaluation. For the privileged teacher, we reimplement TWIST's training procedure, and train on their dataset injected with real-world motion capture data. We train for $1.5$B environment steps on a single NVIDIA RTX 4090, and verify convergence before running evaluations.

\minisection{BeyondMimic} BeyondMimic uses an adaptive sampling strategy and a single set of hyperparameters, trained on individual motion segments. We reimplement BeyondMimic's training procedure in our codebase. We adapt their tracking reward formulation and train on our AMASS dataset. During training, we find that adaptive sampling struggles to converge at-scale so we use a uniform sampler. We train for $1.5$B environment steps on a single NVIDIA RTX 4090, and verify convergence before running evaluations.

\section{Dataset Details}
\label{supp:our:datasets}

\subsection{AMASS}
\textit{AMASS} is a large-scale motion capture dataset unifying data from 15 different sources under a single representation. The data consists of SMPL poses spanning 346 subjects and 11,451 motions across a diverse range of activities, including locomotion, sports, dance, martial arts, and everyday actions. We first filter out infeasible motions that involve object (or environment, such as stairs) interaction. We then use PHC to retarget the human representation to the robot morphology, and further filter motions with sufficiently high retargeting error. We retain 8,123 motions in the robot morphology, using a 90/10\% split for training and evaluation.

\begin{figure*}[t]
    \centering
    \includegraphics[width=\linewidth]{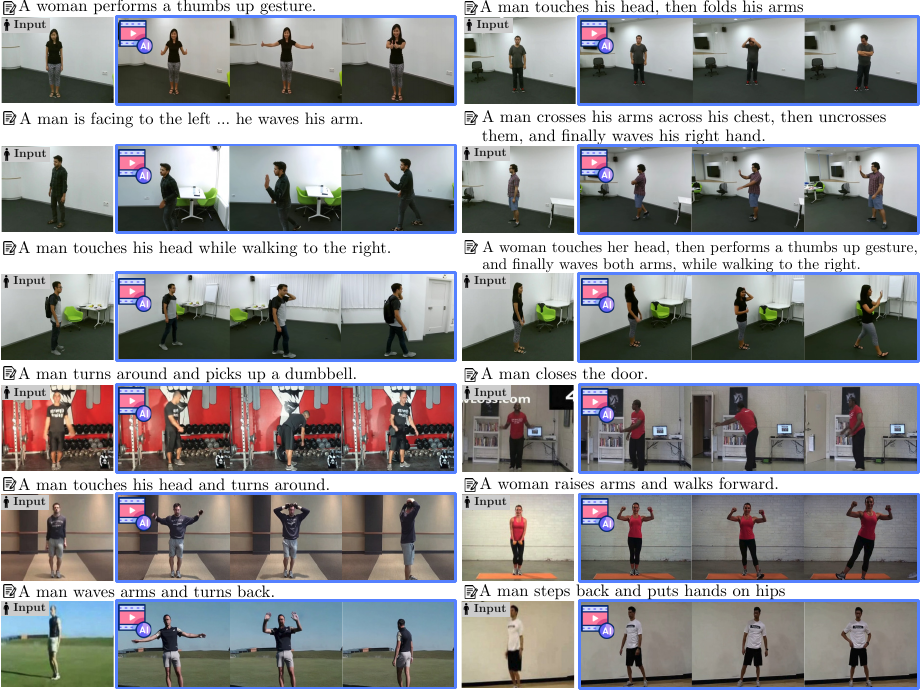}
    \caption{\textbf{Additional Examples of \ourbench.} 
    Our synthetic human-motion dataset is generated using the Wan2.1 and Cosmos-Predict2 video generation models. These videos are produced conditioned on an initial frame and a text prompt describing the action. The dataset spans diverse subjects, environments, and action types including simple gestures and motion compositions.
    }
    \label{fig:supp_dataset_figure}
\end{figure*}

\subsection{\ourbench Details}

This section expands on the construction and characteristics of \ourbench, our synthetic human-motion dataset used to assess zero-shot humanoid control of actions from generated videos. 

\subsubsection{Action Taxonomy}
As shown in \figrref{fig:supp_dataset_figure}, \ourbench spans a diverse set of human motions, categorized by complexity as follows:

\minisection{Simple Upper-Body Motions} These actions involve minimal global body displacement and simple gestures: \textit{Touch head, thumbs up, wave arms}. 

\minisection{Simple Upper-Body Motion+Locomotion} These motions combine periodic walking with upper-body actions: \textit{No gesture + walking, touch head + walking, thumbs up + walking, wave arms + walking}. 

\minisection{Composite Upper-Body Motions} 
These involve multi-step action sequences which chain together simple actions or simultaneous combinations of simple actions: \textit{Touch head $\rightarrow$ thumbs up $\rightarrow$ wave arms; touch head $\rightarrow$ fold arms; raise right hand and point forward $\rightarrow$ fold arms; cross arms $\rightarrow$ uncross $\rightarrow$ wave right hand}.

\minisection{Composite Motions + Locomotion}
These motions combine the multi-step action sequences above with locomotion, such as walking or stepping.

\minisection{In-the-Wild Behaviors}
For in-the-wild behaviors, we condition on 8 frames and 8 subjects from the PennAction dataset to generate a wide range of videos using Cosmos-Predict2. We include (a) simple gestures with and without locomotion, (b) object interaction tasks such as opening doors, picking books, and lifting heavy objects, and (c) action sequences such as walking towards and grabbing an item from a shelf. 

These action categories collectively provide a wide spectrum of motion complexity, from stable and controlled gestures to multi-step behaviors in different views and contexts.

\subsubsection{Video Generation Setup}
\minisection{Text Prompt Design} 
Each video is generated from a text prompt that (1) explicitly describes the target action or composite sequence, (2) avoids stylistic or emotional descriptions to maintain consistency, (3) constrains the environment when beneficial (\eg \textit{in an indoor room}, \textit{walking forward}), and (4) encourages realistic human kinematics (\textit{natural human motion}, \textit{smooth transitions}). 

\minisection{Visual Conditioning Frames}
For both generative models, each clip is conditioned on a single reference frame that determines (1) subject identity including appearance, body proportions and clothing, (2) the surrounding background and scene context, and (3) a camera viewpoint. For Wan2.1, conditioning frames come from synchronized frames in NTU RGB+D videos (in front, left and right views), enabling multi-view setup for the same subject. For Cosmos-Predict2, conditioning frames come from PennAction video frames that represent YouTube-style scenes with natural clutter and varied scene layouts. We select frames in which the subject is upright, in a neutral stance and fully visible without major occlusions to ensure stable initialization for subsequent video generation. 

\minisection{Video Characteristics}
The Wan2.1 videos have a frame resolution of 832$\times$480, a frame rate of 16 fps, and a duration of 5.0s. 
The Cosmos-Predict2 videos have a frame resolution of 768$\times$432, a frame rate of 16 fps, and a duration of 5.8s. 

\begin{figure*}[t]
    \centering
    \includegraphics[width=\linewidth]{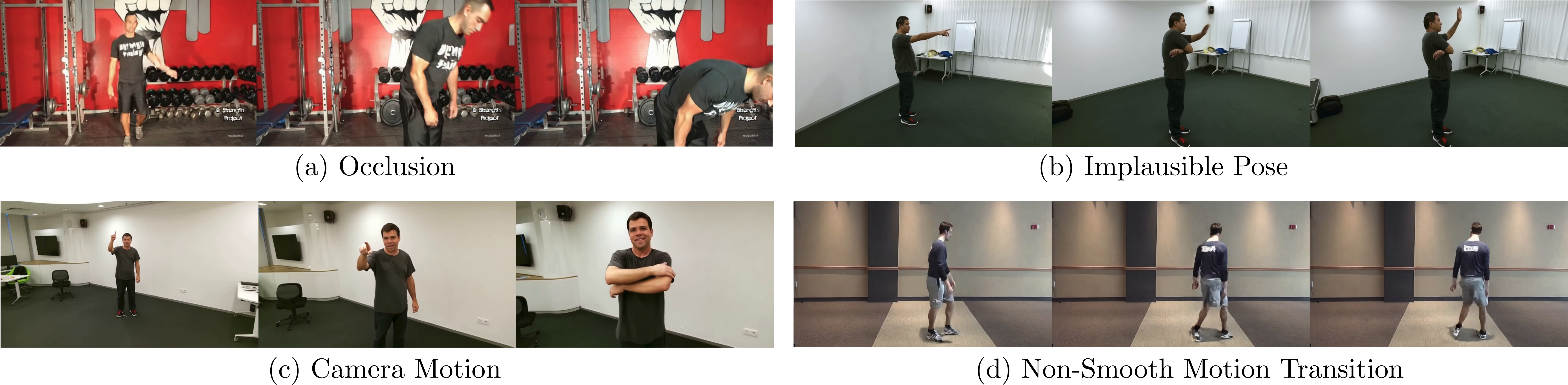}
    \caption{\textbf{Examples of noisy videos in \ourbench}. (a) Occlusion: partial-body visibility and object interference hinder reliable 4D reconstruction. (b) Physically implausible pose: impossible joint configurations (\eg, incorrectly folded arms) produce kinematically invalid reference trajectories. (c) Camera motion: strong viewpoint drift and scene jitter degrade temporal consistency and reduce reconstruction fidelity. (d) Non-smooth motion transition: inconsistent temporal changes (\eg, left/right leg swaps) create discontinuities in the target motion. These artifacts illustrate the noisy, unstable, and sometimes infeasible motion references that humanoid policies must tolerate during zero-shot tracking of generated videos.
    }
    \label{fig:supp_noisy_video_samples}
\end{figure*}

\subsubsection{The challenges in \ourbench}
\ourbench presents a number of challenges for humanoid control. These arise from imperfections inherent in current video generative models, which are amplified during 4D reconstruction and retargeting. Several representative failure modes from these generative artifacts are illustrated in~\figrref{fig:supp_noisy_video_samples}, including partial-body occlusions, physically implausible poses, heavy camera motion, and non-smooth temporal transitions. We summarize the key sources of difficulty below.

\minisection{Appearance and Lighting Drift}
Generated sequences could exihibit subtle changes in subject appearance (\eg, texture details, subject faces, limb proportions), as well as fluctuations in lighting and background. These inconsistencies propagate to noisy or drifting keypoints during 4D lifting, requiring policies to track unstable motion references. 

\minisection{Non-Smooth or Unnatural Motion}
Models can produce abrupt transitions between actions, unrealistic acceleration, or temporally inconsistent motion styles. Such artifacts appear as discontinuous trajectories which can cause stability issues for a humanoid controller.

\minisection{Physically Implausible Poses}
Occasional violations of human kinematics arise, including hyperextension, foot sliding, or brief self-intersections. The policy must learn to tolerate these kinematically infeasible tracking targets. 

\minisection{Occlusions and Camera Effects}
Some video sequences introduce partial-body occlusions, and a significant proportion of videos contain non-static or drifting camera motion. When combined with subtle scene artifacts, these factors reduce 4D reconstruction quality and produce ambiguous trajectories, especially when combined with locomotion.

\begin{figure}[t]
\centering
    \includegraphics[width=\linewidth]{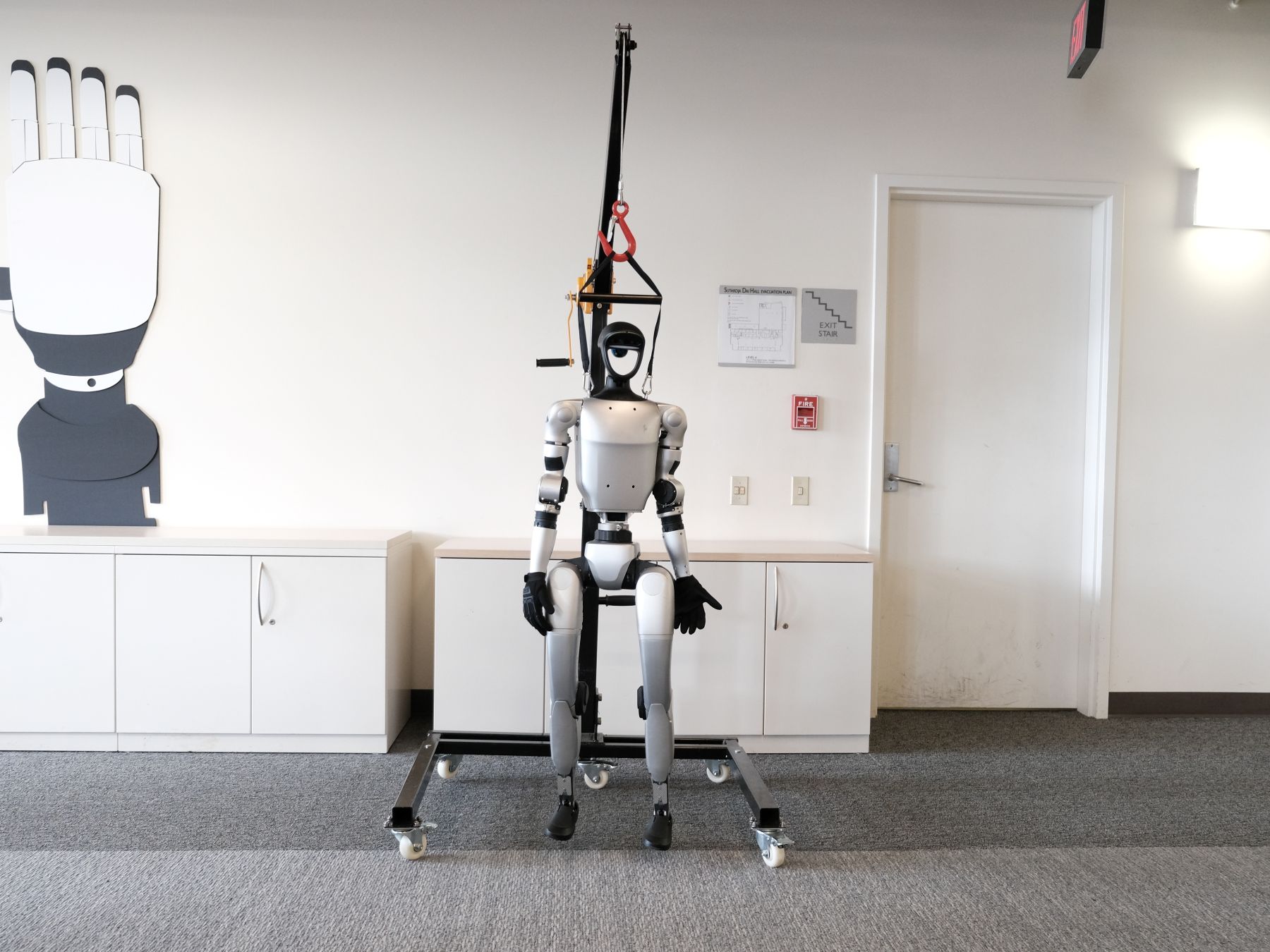}
    \caption{\textbf{Physical Hardware Setup.}}
    \label{fig:supp:hardware}
\end{figure}

\section{Real-World G1 Experiments}
\label{supp:g1}

We use a 23-DoF Unitree G1 humanoid robot equipped with 12 lower body joints, 1 torso joint, and 10 upper body joints for real world experiments. We implement deployment code in Python using the Unitree SDK 2, which runs the policy at 50 Hz on a laptop and sends commands to the onboard PD controler that drives the motors at 500 Hz. We use the default PD gains provided by Unitree, which use $K_p = 200, K_d = 5$ for the hips, $K_p = 300, K_d = 6$ for the knee, $K_p = 40, K_d = 2$ for the ankle, and $K_p = 100, K_d = 2$ for the entire upper body. For safety considerations, especially with noisy input, we keep the policy anchored to a gantry during deployment. We verify all motions in simulation prior to real deployment. See~\figrref{fig:supp:hardware} for a picture of the physical setup.

\end{document}